# Arguments for Nested Patterns in Neural Ensembles

Kieran Greer, Distributed Computing Systems, Belfast, UK.
http://distributedcomputingsystems.co.uk
Version 1.1.

*Abstract* - This paper describes a relatively simple way of allowing a brain model to self-organise its concept patterns through nested structures. Time is a key element and a simulator would be able to show how patterns may form and then fire in sequence, as part of a search or thought process. It uses a very simple equation to show how the inhibitors in particular, can switch off certain areas, to allow other areas to become the prominent ones and thereby define the current brain state. This allows for a small amount of control over what appears to be a chaotic structure inside of the brain. It is attractive because it is still mostly mechanical and therefore can be added as an automatic process, or the modelling of that. The paper also describes how the nested pattern structure can be used as a basic counting mechanism.

## 1   Introduction

This paper[1] describes a relatively simple way of allowing a brain model to self-organise its concept patterns through nested structures. Time is a key element and a simulator would be able to show how patterns may form and then fire in sequence, as part of a search or thought process. The equation that is tested is possibly a simplified version of existing ones ([10] equation 1, for example) which would also consider the synaptic connections in detail. This model is very much generalised and considers the pattern firing properties only. As such, it would be a quicker algorithm to use and so it may allow for more economic test runs that are concerned with this aspect in particular. The algorithm is attractive because it is still mostly mechanical and therefore can be added as an automatic process to any simulator. The process can also be used as a basic counting mechanism and so might be useful for more mathematical operations.

---

[1] accepted for publication at The Science and Information Conference (SAI'14), London, 27 – 29 August, 2014.

The pattern model that is suggested is one of the most commonly occurring models that we have in the real world and so it is clearly understood and could be added to a computer program relatively easily. It is however a very high-level idea, without exact cell workings or connections, for example. After the model and the reasons for suggesting it are described, some tests based on a relatively simple equation will be presented, to show the correctness of the idea.

The rest of this paper is structured as follows: section 2 describes some related work. Section 3 describes the main ideas of the new model. Section 4 describes some tests and results that confirm the main idea, while section 5 gives some conclusions on the work.

## 2    Related Work

This section is more biologically-oriented, where the author is not particularly expert, but the papers might make some relevant points. The aim is to show that the proposed structure is at least practical. The paper [10] is quite closely related and includes a number of important statements. It gives one example of an equation for the firing rate that includes the whole range of inputs, including external sensory and other neurons' excitatory/inhibitory input. It states that for the firing to be sustained, that is, to be relevant, requires sufficient feedback from the firing neurons, to maintain the level of excitation. Once the process starts however, this can then excite and bring in other neurons, when inhibitory inputs also need to fire, to stop the process from saturating. A weighted equation is given to describe how the process can self-stabilise if 'enough' inhibitory inputs fire and a comparison with the equation is given in section 4. The paper [7] also studies the real biological brain and in particular, the chemospecific steering and aligning process for synaptic connections. It notes that there are different types of neuron, synaptic growth and also theories about the processes. While current theory suggests that growth is driven by the neuron itself, that paper would require it to be driven almost completely by the charged 'signal'. Current theory also suggests that the neuron is required first, before the synapses can grow to it. However, they do note a pairwise chemospecific signalling process, as opposed to something that is just random and they also note that their result is consistent with the known preferences of different types of 'interneurons' to form synapses on specific domains of nearby neurons.

The paper [11] also describes how neurons can change states and start firing at different rates. The paper [8] describes that there are both positive and negative regulators. The positive regulators can give rise to the growth of new synaptic connections and this can also form memories. There are also memory suppressors, to ensure that only salient features are learned. Long-term memory endures by virtue of the growth of new synaptic connections, a structural change that parallels the duration of the behavioural memory. As the memory fades, the connections retract over time. So, there appears to be constructive synaptic processes and these can form memory structures. The paper [9] is more computer-based and describes tests that show how varying the refractory (neuron dynamics) time with relation to link time delays (signal) can vary the transition states. They note that it is required to only change the properties of a small number of driver nodes, which have more input connections than others. These nodes can control synchronization locally and they note that depending on the time scale of the nodes, some links are dynamically pruned, leading to a new effective topology with altered synchronization patterns. The structures tested are larger control loops, but it is interesting that the tests use very definite circular pattern shapes. The work of Santiago Ramón y Cajal[2] has been suggested as relevant, in particular, with relation to pacemaker cells. This is definitely interesting and will be discussed in the conclusions of section 5. However, while Cajal appears to classify neurons, based on location defining their function, this paper does not consider different neuron types. It is only interested in location for allowing them to operate as part of a thinking process.

The author has proposed neural network or cognitive models previously [2][3][4][5] and it is hoped that this paper does not contradict that work. The aim has been to build a computer model that copies the brain processes as closely as possible, so as to realise a better or more realistic AI model. It is still a computer program however and a close inspection of how the biological components work has not been considered yet. The goal is to try and make the underlying processes as mechanical, or automatic, as possible, so that the minimum amount of additional intelligence is required for them to work. Earlier themes included dynamic or more chaotic linking, time-based events, pattern formation with state changes, clustering and even hierarchical structures with terminating states or nodes. This paper is associated with some of the earlier work, including the more chaotic neural network structures [2][3], or the pattern forming levels of the cognitive model [4][5].

---

[2] http://www.scholarpedia.org/article/Santiago_Ramón_y_Cajal.

## 3    Reasons For the Firing Patterns Model

It is important to remember that an energy supply is required to cause the neurons to fire. It is probably correct to think that the brain must receive a constant supply of energy to work. If a neuron fires, this would necessarily use up some of the energy, which is why the supply must be refreshed. If thinking about the single neuron, it is thought that ion channels cause the neuron activation, where pressure or force is not the main mechanism[3]. A neuron itself does not have the intelligence to fire, in the sense that it is reactive and not proactive. The fact that inhibitors are used to suppress the firing rate shows that the neurons cannot decide this for themselves. They also need an automatic mechanism to switch off. The activation might be traced back to the external stimulus, which is a continuous energy source, although pressure would be another one [12]. Note also that the brain would be expected to give feedback, which in turn might change the input, and so on. Therefore, if considering the energy used by the system, it would make sense to nest sub-concepts, based on the idea of distance alone.

### 3.1    Sub-Concepts

When thinking about brain firing patterns, it appears to be very random and complex. Pictures or scans of activity however usually show distinct brain areas that are active, where this in itself is interesting. If the firing activity was completely random, then specific areas should possibly not be present, as synapses would travel in any direction to other neurons. So there is an indication here that the firing activity is contained. This then means that it could be inwards, or inside the originally activated area. This can sometimes be almost the whole brain, however. A simple example might also illustrate something. Thinking about a coffee cup, the cup itself can be imagined, sitting by itself. To expand the scenario, possibly a kettle is imagined, filling the cup with water. A specific action would also be invoked here. But is it the case that when the kettle is imagined, the whole kitchen scenario is retrieved? Even if this is done sub-consciously, it would put the scenario into a familiar context, when the kettle and subsequent actions are then easy. This is therefore interesting, because it suggests that the larger activity areas could be these larger concepts that then contain lots of smaller ones inside of them. The kitchen can contain a kettle and a cup, for example. If firing is restricted to the kitchen scenario, it is easy to imagine that it might activate the other smaller kitchen-related concepts. This is also a part of how we try to model the real world, mathematically or formally, in our processes or diagrams, for example.

---

[3] Pressure is not very relevant for this paper, but was used as part of an earlier argument [2] to help the synaptic structures to grow and re-balance.

Further, a single concept can be imagined by itself and even without a background. But the addition of context, invoked by an action or other object, forces the relevant background, even if it is relatively weak. So is the coffee cup and kettle driving the activation and triggering the kitchen that lives somewhere else, or is the span or area of activation now wide enough to activate part of the parent kitchen concept? If the firing was always inwards then activating a larger concept would be difficult, so at least some lateral or outwards positive activation is required. But then again, a coffee cup or kettle might be terminal states that are accessed directly (see possibly the θ value of equation 1 in [10]). As separate pattern groups also need to link, lateral signals could excite a general area between them as well. An action might even originate in a different brain region, bringing all of the connected areas into play. Figure 1 is a schematic of the general idea. There is a larger pattern with nested ones and some excitatory and some inhibitory signals. Traversing the larger area would bring in more of the background patterns or images. So the currently firing pattern is what defines the brain state. If there is no other way of controlling this, the ability to switch off the other areas in an automatic manner is required. If the parent provider encapsulates the new or most active next state, then this activity could be through a relatively simple and easy to understand process. The inhibitors will naturally send more negative feedback to their neighbouring environment, thereby weakening the parent signal compared to the new firing pattern. If new areas inside of them then become active, the process can repeat again. The most obvious catch is the fact that lateral linking and activation is always required and also from other brain regions that perform other functions. It is however, still a natural way to self-organise.

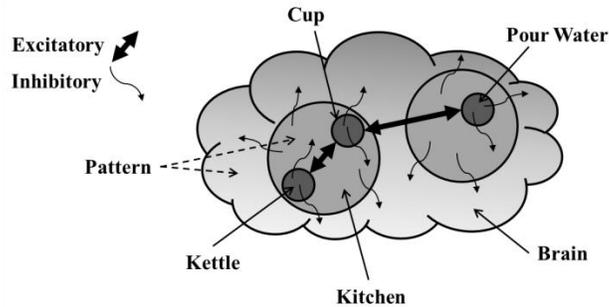

Figure 1. Example of Nested Concepts in a Brain Area.

## 3.2  Mathematical Processes

Another interesting use of the nested patterns is not to retrieve sub-concepts, but to implement a basic timer, counter or even battery, that could be part of more mathematical processes. The idea of battery, counter or timer here, refers to controlling the energy supply of a particular group of neurons. A more general supply is converted into one that can schedule something, or run for a pre-determined amount of time. It might be part of a whole cycle of pattern activations as follows: Some pattern activates another pattern that is the on-switch to a timer or battery. The on-switch activates the outer-most pattern of the nested group that makes up the new structure. This cycles inwards as described, until the inner-most pattern is activated. This might be 3 nested patterns, for example. Each nested pattern, when activated, might send a signal somewhere, but the inner-most one also sends a signal to the off-switch pattern that is beside the on-switch. The off-switch sends inhibitors to the on-switch, asking it to turn off. This then removes the signal inducer to the new structure and the whole cycle can stop.

As this is only an idea, an alternative and possibly better mechanism would be to slowly increase both the excitatory and inhibitory signals. The first activation phase from the outer-most pattern to the first nested one might not activate all neurons one level in. This also means however that they would not all send

DCS                                                                                                                                   29 April 2014

inhibitory signals back. So the outer-most pattern might be able to send several phases of signal before it receives an overwhelming inhibitory signal. The same situation can occur between any of the nested pattern sets. Continual activation signals can switch on more neurons the next level in, but then they also send more inhibitory signals back. If the excitatory signal is mostly inwards and the inhibitory one mostly outwards, this should result in the whole region eventually switching itself off.

Slightly more doubtful: if the inhibitory signal only affects active neurons, then they can possibly fire in any direction, because the inner patterns will receive less than the outer ones, based on time events and so the outer ones will switch off first. So there would still be the desired and gradual build-up of signal and shut-down afterwards. Note that these cases require a constant, external energy supply, which then gets shut-down or ignored. It would also be helpful if inhibitors could change a neuron state without switching it off completely and ideas of localised firing already exist [9]. The schematic of Figure 2 tries to describe the most general case. Some area of the brain excites and starts the outer-most pattern firing. This is the 'on' switch with a signal to the outer-most circle. The pattern cycles through to the inner-most one that can then ask for the provider to switch off. This is the 'off' switch. Each nested pattern can also send a signal somewhere else, which would implement the counting mechanism.

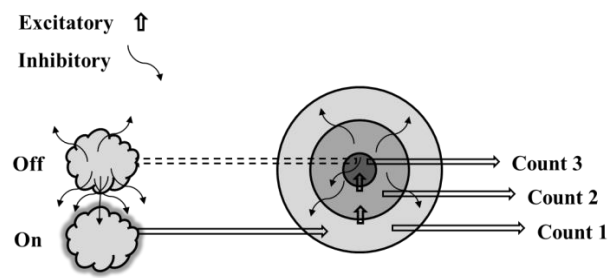

Figure 2. Example of a Timer, Counter or Battery.

The paper [10] notes another model that already exists called Synfire chains. Synfire chains fire in a definite outwards direction and offer some degree of control through firing stages at different levels. This then leads to problems of explosion from a sustained input and requires noise or other to control the firing rates.

So the main question for the model of this paper is whether it can actually occur naturally, as other formations appear to be outward facing. It is worth remembering that clear pattern boundaries get created however.

As well as deciding to fire, this paper would require the neuron to intelligently control direction. Why would the neurons prefer to fire inwards instead of any direction? The theory of this paper however allows that intelligence to be replaced with an economic reason, based on the conservation of energy. If thinking about stigmergic systems [1] for example, the ant colony selects the most economic path unintentionally and neurons equally influence each other. The idea of grouping more closely, neurons that fire at the same time, is also the well-known doctrine of Hebb [6]. The search process would also conceivably converge on terminal states [2], where Figure 3 could help to describe the economic argument. The idea of 'neurons that fire together wire together' requires a link between the two or more groups involved. If search occurs from a broad group to a smaller terminal state; then if that search is outwards, as in Figure 3a, the distance between the terminal states and the nodes in general is greater than if it is inwards, as in Figure 3b. Note in particular the case where the terminal states join to complete a circuit. Also, exactly as with stigmergy, if both pattern sets receive the same amount of energy, Figure 3b will reinforce more, because the signal can take a shorter route. That might just provide a reason why it is easier for the inward pointing search to then connect with another related search area, than the outward pointing one. Therefore, even by chance, a random process might prefer the inward facing groups.

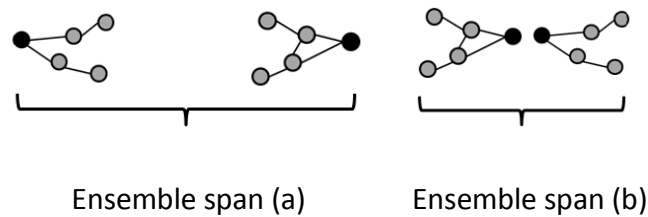

Ensemble span (a)       Ensemble span (b)

Figure 3. The two search process that ends with closer terminal states (b) is more economic.

## 4 Testing and Results

Testing of the theory can be carried out by implementing some basic reinforcement algorithms and updating specific node values, to simulate the different timings of the node pattern activations. The traditional increment/decrement reinforcement algorithm worked well enough to give the desired result. With that, the node value is incremented with excitatory input and decremented with inhibitory input. The decrement value can be weighted to be only a fraction of the increment. Some assumptions are made with regards to the neurons, which helps to simplify the problem further:

- Each neuron has only one excitatory output and one inhibitory output.
- The excitatory output goes only to the other neurons firing in the same pattern.
- The inhibitory output goes only to any other neuron that is in any parent pattern.
- Neurons are in the same pattern if they fire at the same time. This is measured by time increments $t_1$ … $t_n$.

### 4.1 Test Conditions

The equation 1 for firing rate networks in [10] is probably a complete version of the equation that might be used. These tests only consider the excitatory/inhibitory part, to measure how the patterns will switch on and off through their interactions. The firing interactions are further restricted by the aforementioned assumptions. The resulting test equation for this paper could therefore be written as follows:

$$X_{it} = \sum_{p=1}^{P_i} Ept - (\sum_{k=Pj}^{l} \sum_{y=1}^{m} \sum_{j=1}^{n} (Hjy * \delta))$$

where $y \neq t$ and $i \in P_i$ and $not\ j \in P_i$, and

$X_{it}$ = total input signal for neuron i at time t.
$E_p$ = total excitatory input signal for neuron p in pattern P.
$H_{jy}$ = total inhibitory input signal for neuron j at time y.
$\delta$ = weights inhibitory signal.
t = time interval for firing neuron.
y = time interval for any other neuron.

DCS                                                                                                                                      29 April 2014n = total number of neurons.

m = total number of time intervals.

l = total number of active patterns.

$P_i$ = pattern for neuron i.

P = total number of patterns.

In words, the tests measure how the total signal input to each neuron pattern changes. All neurons in the same pattern fire at the same time and send each other their positive signal. Any active neuron also receives a negative signal from any other nested pattern neuron. If two nested patterns are active, for example, the inner-most sends inhibitory signals to both the outer-most pattern and the first nested one. The first nested pattern sends inhibitory signals to just the outer-most one. Over time, neurons continue to fire based on - total pattern firing strength minus total inhibitory firing strength from all other nested patterns.

### 4.2    Test Results

The test results are quite straightforward and show the desired set of relative counts or signal strengths. Just the traditional increment/decrement algorithm is shown in Table **1**. There are 25 neurons in total and 5 in each nested pattern. The inhibitory signal is set to be half that of an excitatory one, but if a pattern only contains 5 neurons, that leaves a possible 20 other neurons that might send inhibitory signals. Each firing cycle activates a new nested pattern, until all patterns are active. After that, each firing cycle would update signals from all patterns. The inhibitory signal is sent from the inner pattern to its outer ones only, so the inner-most one does not receive inhibitory signals. When all patterns are active, the inhibitory signal builds up to overwhelm the excitatory signal. This of course, depends on the pre-set relative strengths and numbers of excitatory and inhibitory signals.

Neurons 1 to 5 are the outer-most pattern. Neurons 6 to 10 are the first nested pattern and so on, until neurons 21 to 25 are the inner-most nested pattern. At time t1, the first pattern only fires (neurons 1 to 5). At time t2 pattern 1, then pattern 2 fires. At time t3, pattern 1, then pattern 2, then pattern3 fire, and so on. The outer patterns have more excitatory input to start with, but as the other patterns switch on and send negative feedback, eventually they will switch off the outer patterns. This would then actually starve the inner patterns of input, until they switch off as well.

Table 1. Relative Pattern Strengths after Firing Sequences.

| Neurons | t = 3 | t = 4 | t = 5 |
|---|---|---|---|
| 1 | 7.5 | 5.0 | 0.0 |
| 2 | 7.5 | 5.0 | 0.0 |
| 3 | 7.5 | 5.0 | 0.0 |
| 4 | 7.5 | 5.0 | 0.0 |
| 5 | 7.5 | 5.0 | 0.0 |
| 6 | 7.5 | 7.5 | 5.0 |
| 7 | 7.5 | 7.5 | 5.0 |
| 8 | 7.5 | 7.5 | 5.0 |
| 9 | 7.5 | 7.5 | 5.0 |
| 10 | 7.5 | 7.5 | 5.0 |
| 11 | 5.0 | 7.5 | 7.5 |
| 12 | 5.0 | 7.5 | 7.5 |
| 13 | 5.0 | 7.5 | 7.5 |
| 14 | 5.0 | 7.5 | 7.5 |
| 15 | 5.0 | 7.5 | 7.5 |
| 16 | 0.0 | 5.0 | 7.5 |
| 17 | 0.0 | 5.0 | 7.5 |
| 18 | 0.0 | 5.0 | 7.5 |
| 19 | 0.0 | 5.0 | 7.5 |
| 20 | 0.0 | 5.0 | 7.5 |
| 21 | 0.0 | 0.0 | 5.0 |
| 22 | 0.0 | 0.0 | 5.0 |
| 23 | 0.0 | 0.0 | 5.0 |
| 24 | 0.0 | 0.0 | 5.0 |
| 25 | 0.0 | 0.0 | 5.0 |

## 5    Conclusions

The purpose of this paper is to show how nested, or more specific patterns, may become the main focus in a generally excited area. They might even be used as part of more complex mathematical processes. Rather than the exact details of how they might be created, or link to each other, etc., the paper describes how they might be useful. Simulation would be easier if the interaction between the patterns only was considered, using a general equation for their relative strengths. The mechanical process can also work with a minimum of complexity and would allow these patterns to form and fire in sequence. It would also realise some level of natural order, which would be better than the very random and chaotic structures that appear to be present. This type of natural ordering would help to simplify the self-organisation process, making it more economic. As described in section 2, the mechanism is still compatible with earlier work.

The main conclusions from the theory are: 1. Nested patterns can offer a level of control over firing sequences. 2. They can also be used as a mathematical mechanism, because the nesting makes it local, finite and timed. Some test results show how easy it is to implement the basic algorithm and that it obviously works as defined. The main test conclusions are probably: 1. Neurons firing together in a pattern should receive fewer inhibitory signals asking them to shut down. 2. If nested patterns are used, then a child pattern can ask its parent to shut down more easily and thereby gain a prominent state itself.

There also needs to be a forced method to access any concept directly, or to join up separate patterns or groups, possibly over long distances, but that is already known and part of a different problem. It is still unclear if this is a realistic scenario, even if it appears to be a practical one. The structure in this paper has not been refuted previously, as far as the author is aware. If considering the energy used by the system, it would even make sense to nest sub-concepts, based on distance alone. Also, if considering the human evolution, it is plausible to consider how a visual mechanism that provided self-organisation was changed slightly to produce a more calculating mechanism that used the same underlying architecture. It is a believable step for how we may have become more intelligent, by being able to count and manage more complex operations through the timed control of firing events. Timing has already been demonstrated by Cajal and the pacemaker cells that control the intestinal tract. As far as the brain is concerned however, while this work argues for mechanical processes, we then need to add the 'intelligent' overriding control again. If our consciousness is the repeated firing of our memories and images that we store, in order for this to be sane, probably requires that it is not always random. While the new mechanism can control very localised sequences, it does not control when a localised sequence, relating to something, starts to fire. Unless that is again, just a very weak stimulus that needs to be satisfied.

## Acknowledgment

Thank you to The Science and Information Organisation for financial assistance with relation to the conference fees.